\documentclass[]{article}
\usepackage[letterpaper]{geometry}
\usepackage{mtsummit2017}
\usepackage{times}
\usepackage{url}
\usepackage{latexsym}
\usepackage{natbib}
\usepackage{layout}

\usepackage{multirow}
\usepackage{arabtex}
\usepackage{graphicx}
\usepackage{tabularx}
\usepackage{expex}
\usepackage{siunitx}
\usepackage{booktabs}
\usepackage{wrapfig}

\newcommand{\hide}[1]{}
\usepackage[T1]{fontenc}

\newcommand{\AHAMZAUP}{{\^{A}}}

\newcommand{\TAMARBUTA}{{$\hbar$}}

\newcommand{\DHA}{{\dh}}
\newcommand{\SHIN}{{\v{s}}}
\newcommand{\ZA}{{\v{D}}}
\newcommand{\AYN}{{$\varsigma$}}

\begin{document}

\title{Low Resourced Machine Translation\\ via Morpho-syntactic Modeling:\\ The Case of Dialectal Arabic}

\author{\hfill \name{\bf Anonymous Submission} \hfill} 

\author{\name{\bf Alexander Erdmann} \hfill  \addr{ae1541@nyu.edu}\\ 
\addr{Computational Approaches to Modeling Language (CAMeL) Lab,\\ New York University Abu Dhabi\\Department of Linguistics, Ohio State University}\\
\AND
        \name{\bf Nizar Habash} \hfill \addr{nizar.habash@nyu.edu}\\ 
        \addr{Computational Approaches to Modeling Language (CAMeL) Lab,\\ New York University Abu Dhabi}\\
\AND
        \name{\bf Dima Taji} \hfill \addr{dima.taji@nyu.edu}\\ 
        \addr{Computational Approaches to Modeling Language (CAMeL) Lab,\\ New York University Abu Dhabi}\\
\AND
       \name{\bf Houda Bouamor} \hfill \addr{hbouamor@cmu.edu}\\
        \addr{Department of Computer Science, Carnegie Mellon University Qatar}
}

\maketitle
\setarab
\novocalize

\pagestyle{empty}

\begin{abstract}
\end{quote}

We present the second ever evaluated Arabic dialect-to-dialect machine translation effort, and the first to leverage external resources beyond a small parallel corpus. The subject has not previously received serious attention due to lack of naturally occurring parallel data; yet its importance is evidenced by dialectal Arabic's wide usage and breadth of inter-dialect variation, comparable to that of Romance languages. Our results suggest that modeling morphology and syntax significantly improves dialect-to-dialect translation, though optimizing such data-sparse models requires consideration of the linguistic differences between dialects and the nature of available data and resources. On a single-reference blind test set where untranslated input scores 6.5 BLEU and a model trained only on parallel data reaches 14.6, pivot techniques and morpho-syntactic modeling significantly improve performance to 17.5.
\end{abstract}

\section{Introduction} \label{intro}

Arabic is widely spoken and highly diglossic, with Modern Standard Arabic (MSA) representing the high register shared across the Arab World in educated circles. In contrast, the many spoken dialectal Arabic varieties (DA) are somewhat if not entirely mutually unintelligible, e.g., Moroccan and Kuwaiti. \cite{ChiangEtAl:2006} compare the linguistic variation among Arabic dialects to that among Romance languages, indicating the need for machine translation (MT) between these dialects. However, while much MT research has been devoted to translating between Romance languages~\citep{corbi2005open2,armentano2006open,Koehn_09a_MTS}, we are aware of only one work on Arabic DA-to-DA MT~\citep{meftouh2015machine}. It deals mainly with Maghrebi dialects and utilizes only a small parallel corpus.\footnote{Maghrebi dialects are those spoken in Morocco, Algeria, Tunisia and Libya.} This work focuses on the Egyptian and Levantine dialects, leveraging various available resources such as a morphological analyzer and additional monolingual and multilingual data.\footnote{Levantine covers the dialects spoken in Lebanon, Syria, Palestine and Jordan.} Compared to other dialects, Egyptian and Levantine's wider range of available data/resources allows us to evaluate more MT approaches using different combinations of these data/resources. Thus, in future work on DA pairs which may not have the same data/resources, we can tailor MT systems based on this paper's findings.


The main challenge in developing DA-to-DA MT systems is the lack of data. While many Romance languages are official languages with written standards, naturally occurring in parallel corpora like the European Parliament \citep{koehn2005europarl}, DA has no official status and was rarely written until the advent of social media.\footnote{Recently, two parallel Arabic translations were created for 12,000 sentences from the European Parliamentary proceedings, but both are in MSA \citep{habash2017parallel}.} The recent release of the first parallel multi-dialectal corpora~\citep{bouamor2014multidialectal,meftouh2015machine} has enabled seminal, albeit low-resource MT experiments. We present some shortcomings of these corpora and introduce an in-house, under-development corpus. Then we explore different means of leveraging external resources, e.g., Egyptian-to-English and Levantine-to-English data and an Egyptian tokenizer and morphological analyzer \citep{CALIMA:2012,ARZTB:2014,MADAMIRA:2014}. We conduct experiments in a range of data-sparse settings and show the effect of morpho-syntactic features on the DA-to-DA MT performance. Our approach can be extended to other DA pairs and other closely related languages and dialects \citep{tyres2017rulebased}. 




\section{Related Work} \label{related-work} 
An increasing amount of research has been conducted on dialectal Arabic NLP; however, most dialectal MT efforts translate from DA to MSA or English. The only other DA-to-DA work we are aware of focuses on manipulating language model smoothing parameters to optimize data sparse MT performance \citep{meftouh2015machine}.

\subsection{Dialectal Arabic Machine Translation}
\label{DAMT}
While only \cite{meftouh2015machine}
have evaluated DA-to-DA MT, many others have addressed MT between DA and other languages. 
\cite{zbib2012machine} attempted to translate from Egyptian and Levantine to English and found that pivoting through better resourced MSA was not useful due to register and domain differences. MSA, the higher register, is rarely used to discuss the day-to-day matters frequently treated with DA, causing a domain mismatch. However, several approaches have since presented alternative results \citep{Sawaf:2010,salloum2011dialectal,Salloum+Habash:2012,sajjad-darwish-belinkov:2013:Short,durrani2014improving}. These use rule-based or hybrid methods to identify mappings from DA to MSA before translating to a target language (usually English).
Additionally, \cite{tachicart2014hybrid} report results on adapting an approach designed for MSA to Moroccan translation to translate in the inverse direction (Moroccan to MSA).

\subsection{Dialectal Arabic Data}
\label{data}

Several newly developed corpora have facilitated the recent surge in dialectal NLP work. 

The DARPA BOLT (Broad Operational Language Translation) project sponsored the creation of a large number of resources,\footnote{Pointers to the Linguistic Data Consortium's BOLT resources can be found here: \url{https://www.ldc.upenn.edu/collaborations/current-projects/bolt}.} including a sizeable data set of DA sentences paired with their English translations. This data set consists of 2.2 million words of Egyptian and 1.5 million words of Levantine which were harvested from SMS messages and online sources like weblogs before being translated.

As for monolingual corpora, \cite{zaidan-callisonburch:2011:ACL-HLT2011}'s Arabic Online Commentary (AOC) corpus contains 52 million words of mixed MSA and DA from news articles and readers' comments. \cite{cotterell2014multi} add modest amounts of Twitter data to this corpus, though we find the domain difference harmful for language modeling and drop it in our experiments.
\cite{khalifa2016large}'s GUMAR corpus contains over 100 million words of Gulf Arabic and a smattering of other dialects, all taken from internet novels, a genre of long conversational novels shared anonymously on online forums popular among female teenagers. Other monolingual DA corpora like Tunisiya \citep{mcneil2011tunisian}, the Curras corpus of Palestinian-Levantine \citep{jarrar2014building}, and those corpora presented in \cite{al2016morphologically}, focus on different dialects or are too small to be relevant to Egyptian-to-Levantine MT.



As for DA-to-DA data, \cite{bouamor2014multidialectal} present the first corpus with 2,000 7-way parallel sentences of Egyptian, Tunisian, three Levantine dialects (Syrian, Jordanian, Palestinian), MSA, and English, all translated from Egyptian sentences harvested from the web. The authors concede that many Levantine sentences seem to be influenced by the Egyptian, likely because translators were primed with Egyptian expressions they might understand, but would not produce naturally.
The same concern applies to the 6,400 sentence, 6-way parallel PADIC corpus used in \cite{meftouh2015machine}, as all translations were derived from DA or MSA. 
When developing the 12,000 sentence multi-dialectal corpus used in our experiments, we avoided such priming effects by asking translators to produce translations starting from English sentences taken from the Basic Travel Expressions Corpus (BTEC)~\citep{takezawa2002toward}. 


Other relevant resources include AVIA,\footnote{\url{http://www.umventures.org/technologies/arabic-variant-identification-aid-avia}} a small but rich multi-dialect reference grammar with contextual examples, and Tharwa \citep{Tharwa2012}, a 4-way English, MSA, Egyptian, Levantine lexicon with rich linguistic annotation.

\subsection{Pivot Machine Translation}
Pivoting is an MT technique used to combat data sparsity when more source-to-pivot and pivot-to-target data is available than source-to-target parallel data \citep{pivot,hajic:etal:2004,wu-wang:2007:ACLMain,habash-hu:2009:WMT}. In this work, we use a specific form of pivoting: phrase pivoting. This involves aligning source-to-pivot and pivot-to-target data, extracting pairs of phrases into two phrase tables, then combining them into a single source-to-target phrase table based on shared pivot phrases \citep{Utiyama:2007}.

Our work is similar to \cite{el2013language}, who use English to translate from Persian to Arabic via phrase pivoting. They introduce connectivity strength constraints to weight learned Persian-to-Arabic phrase-pairs in the table by considering how well each pair can be aligned through an English pivot phrase (discussed further in Sections \ref{baseline} and \ref{morph}).
In follow-up work, \cite{kholy2015morphological} add morphological constraints for translating related, morphologically rich languages Arabic and Hebrew, via morphologically-poor English. These constraints help preserve fine grained morphological distinctions like gender agreement which cannot otherwise be accurately translated via a morphologically poor pivot that does not make such distinctions, i.e., English.

\section{Data Preparation} \label{preprocessing} 
\label{DAPP}
All data used in our experiments comes from sources mentioned in Section \ref{data}. As displayed in Table \ref{dataTable}, we split our 12,000 sentence BTEC parallel corpus into training, tuning, dev, and blind test sets, which are constant across all experiments. Also, in some experiments, we use additional monolingual and pivot data from AOC and the BOLT corpus, respectively.


\begin{table}[tbh]
\setlength{\tabcolsep}{3pt}
\centering
\begin{tabular}{|c|c|c|c|}
\hline
\textbf{Data Set} & \textbf{Dialect} & \textbf{Description} & \textbf{Size} \\ \hline\hline
BTEC-train & Egy--Lev & Parallel & 8,000 \\ \hline
BTEC-tune & Egy--Lev & Parallel & 500 \\ \hline
BTEC-dev & Egy--Lev & Parallel & 1,500 \\ \hline
BTEC-test & Egy--Lev & Parallel & 2,000 \\ \hline \hline
BOLT-egy & Egy--Eng & Pivot & 410,000 \\ \hline
BOLT-lev & Eng--Lev & Pivot & 180,000 \\ \hline \hline
AOC-egy & Egy & Monolingual & 9,000 \\ \hline
AOC-lev & Lev & Monolingual & 5,000 \\ \hline
\end{tabular}
\caption{Data used in all experiments. Size reported in number of sentences.}
\label{dataTable}
\end{table}


Similar to MSA, DA is morphologically and syntactically rich, posing several challenges for MT systems. 
To be able to leverage morpho-syntactic features, we ran our Egyptian and Levantine data through MADAMIRA~\citep{MADAMIRA:2014}, an Arabic morphological analyzer and disambiguator trained for MSA (MADAMIRA-MSA) and Egyptian (MADAMIRA-EGY). 
Unfortunately, the Levantine version of MADAMIRA is still under development~\citep{eskander2016:COLING}, so we use MADAMIRA-EGY to process both our Egyptian and Levantine corpora. \cite{jarrar2014building} and \cite{khalifa2016large} show that using MADAMIRA-EGY to process non-Egyptian DA data yields better results than MADAMIRA-MSA.
To minimize the analyzer's bias towards Egyptian when processing Levantine data, we do not allow it to make orthographic changes.
This limits the effects of misanalyzing many Levantine words, such as <hAl.h.z> \textit{hAlH{\ZA}} `this luck', which can be incorrectly Epgyptian{\it ized} as <.ha'al.h.z> \textit{H{\AHAMZAUP}lH{\ZA}} -- the Egyptian future particle +<.h> \textit{H+} together with an MSA verb <'al.h.z> \textit{{\AHAMZAUP}lH{\ZA}} `I perceive'.\footnote{Arabic transliteration is presented in the Habash-Soudi-Buckwalter scheme \citep{HSB-TRANS:2007}.} 

A number of tokenization and segmentation schemes are available for Arabic \citep{Habash_10_book}. Some separate only punctuation and digits. Others, such as ATB and D3, separate different sets of clitics from the base word. Whereas D3 segments all clitics, ATB leaves attached the definite article, <Al> \textit{Al}.
The optimal segmentation for our task is D3 \citep{sadat2006combination}, as the aggressive tokenization mitigates for data sparsity. Typically, these tokenization schemes involve orthographic rewrite rules to ensure that the base word matches its non-cliticized form to minimize sparsity \citep{el2012orthographic}.
Such rules depend on the morphological template of the word and the clitics attached to it. For a word such as <.hyktbwhA> \textit{HyktbwhA} `and they will write it', the basic D3 tokenization is {\it H+ yktbwA +hA}. The extra {\it A} is added to the base word to minimize sparsity as this is how it would appear if no suffix had been appended.\footnote{In all of the work presented in this paper we apply <Y>/<A> \textit{Alif/Ya} normalization \citep{el2012orthographic}.}

Since we do not have ideal tools for processing (tokenizing and detokenizing) Levantine, we opt for a stricter surface-word-oriented segmentation that guarantees recovering the form by simple concatenation when detokenizing. Thus, for <.hyktbwhA> \textit{HyktbwhA} `and they will write it', the desired D3 segmentation is {\it H+ yktbw +hA}.
This may increase data sparsity slightly, but more importantly, as mentioned previously, this limits the extent to which words can be misanalyzed or overly Egyptian{\it ized}.
To achieve this, we extend a DA morphological database with suffix and prefix segmentations, adding a wrapper on top of MADAMIRA to generate the proper segmentation for each analysis. The database extension is automatic and the segmentation is deterministic, following D3 segmentation rules. 
This allows us to (i) apply this extension to other databases in other dialects that follow the structure of the MADAMIRA database, and (ii) expand our application to dialects that do not have any available analyzers yet.  


\section{Baseline Models} \label{baseline}
We use the phrase-based statistical MT platform, Moses~\citep{koehn2007moses} to build multiple Egyptian-to-Levantine MT systems: one that only trains on parallel data, another that fabricates pseudo-parallel training data from additional monolingual data, and a third model utilizing pivot data through English. While neural MT has been successfully applied to MSA \citep{almahairi2016first}, we opt for statistical MT as data sparsity and other factors render neural techniques impossible for DA \citep{DBLP:journals/corr/ZhangKCS16}.
\cite{luong2015stanford}'s English-to-Vietnamese neural MT system, for instance, leverages 10 times more parallel data then we use in our experiments, yet still fails to outperform a statistical baseline.
Furthermore, their training and testing data is from a single domain with standardized spelling, i.e., limited token:type ratio, which \cite{farajian2017neural} suggest should greatly facilitate neural MT performance. Given our sparsity of DA data and lack of spelling conventions, we can neither rely on homogeneous training/testing domains nor low token:type ratios and must resort to statistical MT.


We evaluate the output of our MT systems via BLEU scores \citep{papineni2002bleu}, comparing them to a single reference in detokenized space.
\textsc{No-Translation}, scoring 6.48, compares the original, unchanged Egyptian input to the Levantine reference. The results as well as data requirements are reported in Table \ref{baselineResults}. 



\begin{table*}[!t]
\setlength{\tabcolsep}{3pt}
\centering
\begin{tabular}{|c|c|c|c|c|c|}
\cline{4-6}
\multicolumn{3}{c|}{} & \multicolumn{3}{c|}{\textbf{Required Data}} \\ \hline
\textbf{Model} & \textbf{BLEU} & \textbf{Out-of-vocabulary} & \textbf{Parallel} & \textbf{Monolingual} & \textbf{Pivot} \\ \hline

\textsc{No-Translation} & 6.48 & N/A & & & \\ \hline
\textsc{Direct} & 15.44 & 4.6 & X & & \\ \hline
\textsc{Synthetic} & 16.75 & 0.8 & X & X & \\ \hline
\textsc{Phrase Pivot} & 6.77 & 1.4 &  &  & X \\ \hline
{\bf \textsc{Dir+PP}} & {\bf 17.41} & 0.9 & X &  & X \\ \hline
\textsc{Synthetic-Dir+PP} & 16.81 & 0.8 & X & X & X \\ \hline

\end{tabular}
\caption{Baseline BLEU scores given different requirements: parallel, monolingual, or pivot data. Out-of-vocabulary rates are presented as percentages for each model.}
\label{baselineResults}
\end{table*}

\subsection{The Direct Model}
The most basic statistical system, the \textsc{Direct} model can be extended to any dialect pair with parallel data. It is trained only on our BTEC parallel corpus, with some additional monolingual data for language modeling. This model leverages a 2.4 million token 5-gram language model trained using KenLM \citep{heafield2011kenlm}, consisting of Levantine data from the AOC corpus, BOLT, and BTEC. 

Following \cite{kholy2015morphological}, we perform word alignment using the grow-diag-final algorithm \citep{och2003systematic} and we restrict the maximum length of extracted phrases to 8 tokens. Our D3 tokenization is slightly more aggressive than \cite{kholy2015morphological} who use ATB, so we experimented with marginal increases in the maximum allowable phrase length but found them to have no significant effects on performance.

As shown in Table~\ref{baselineResults}, this basic model greatly outperforms the \textsc{No-Translation} baseline at 15.44 BLEU, but suffers from a high rate of out-of-vocabulary (OOV) words given that it is only trained on a small amount of parallel data. Furthermore, the model seems to learn noisy weights for many of the phrase pairs it extracts due to the infrequency with which they are encountered during training.

\subsection{The Synthetic Model}
Inspired by \cite{schwenk2009translation}, we use additional monolingual data to build a \textsc{Synthetic} MT system. First, we use the \textsc{Direct} model to translate all of the BOLT Egyptian data to Levantine. Then we build an inverse model identical to the \textsc{Direct} model, but from Levantine to Egyptian, and use it to translate the BOLT Levantine data into Egyptian.
Finally, we learn a new phrase table from our newly generated parallel corpus consisting of the original 8,000 training sentences, 410,000 BOLT Egyptian-to-generated-Levantine sentences, and 180,000 BOLT Levantine-to-generated-Egyptian sentences.

While \cite{schwenk2009translation} implement this technique in a slightly different manner for the purpose of domain adaptation, we use it to reduce noise in the phrase table.
Due to sparsity of parallel data, the \textsc{Direct} model is hard pressed to distinguish good low frequency phrase pairs from bad ones. Adding synthetic data to the model enables it to learn better alignments for low frequency phrase pairs by getting exposure to a variety of different contexts in which such phrases can occur.
This system significantly improves over the \textsc{Direct} model, scoring 16.75 BLEU, representing our best solution for DA-to-DA MT that does not require pivot data.

\subsection{The Phrase Pivot Model}
Following \cite{el2013language}, we use the BOLT data to phrase pivot through English. Phrase pivoting drastically increases vocabulary coverage; however, it also produces a phrase table with many poorly connected phrase pairs as well as phrase pairs which erroneously translate morpho-syntactic features that cannot be conveyed through morphologically-poor English.
The \textsc{Phrase Pivot} model addresses the poor connectivity issue by adding \cite{el2013language}'s connectivity strength constraints.
These identify how many Egyptian and Levantine tokens in a given Egyptian-to-Levantine-via-English phrase pair can be aligned to each other via corresponding alignments to the same English token.


\begin{wrapfigure}{r}{0.5\linewidth}
\centering
 \centering
  \includegraphics[width=4cm]{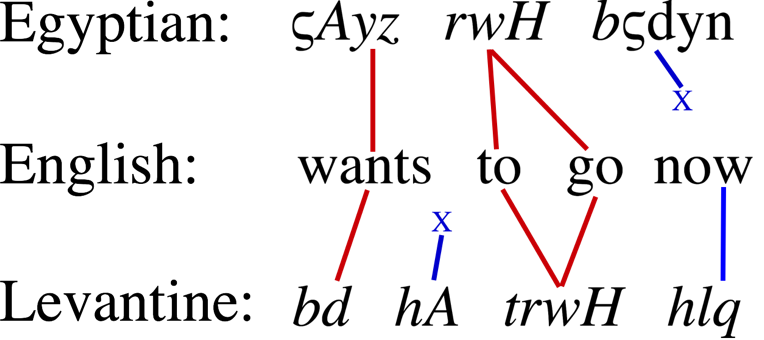}
  \caption{Identifying the connections between an Egyptian phrase and Levantine phrase which were both independently (and noisily) mapped to the same English phrase during pivoting.}
  \label{datFigDo}
\end{wrapfigure}

For example, the noisy Egyptian-to-Levantine phrase pair in Figure \ref{datFigDo}, would receive a connectivity score of 0.75 from the Egyptian side because 3 of the 4 alignments---those to `wants', `to', and `go'---connect through the English pivot phrase to a Levantine token on the other side. The connectivity score from the Levantine side would be 0.6 because 3 of the 5 Levantine alignments connect all the way through. {\it hlq} does not count towards the 3 connections because while it connects to the English token `now', no Egyptian token connects to `now' from the other side.
This example also exhibits the issue that will be addressed in Section \ref{morph}, that morpho-syntactic properties are not accurately conveyed through morphologically deprived English, as {\it {\AYN}Ayz} and {\it bd} connect through `want', though {\it {\AYN}Ayz} implies a masculine subject whereas the suffix of {\it bd}, {\it hA}, entails that the subject of the Levantine sentence is in fact third-person feminine.

This \textsc{Phrase Pivot} model can be extended to any DA pair with pivot data, but only marginally outperforms the \textsc{No-Translation} baseline at 6.77 BLEU. However, used in conjunction with the \textsc{Direct} model, the Direct + Phrase Pivot (\textsc{Dir+PP}) model increases OOV coverage, boosting performance to 17.41 BLEU, almost 2 full BLEU points over the \textsc{Direct} baseline.
We also re-ran the \textsc{Synthetic} model using \textsc{Dir+PP} to fabricate parallel data instead of \textsc{Direct}, however, this did not improve performance. 
It is possible that the types of \textsc{Direct} model errors which are corrected by \textsc{Dir+PP} versus those corrected by \textsc{Synthetic} are similar. Thus, when training on fabricated parallel data, the \textsc{Synthetic-Dir+PP} model may reinforce its own errors more than learn to fix them.

\section{Leveraging Morphology and Syntax} \label{morph}
The best baseline system, \textsc{Dir+PP} still fails to adequately handle Arabic's rich morphology and syntax, as illustrated by Figure~\ref{fig:ex1}, where part-of-speech (POS) is not preserved in the output. 
A minimally different correct version of the example in Figure~\ref{fig:ex1} would simply replace verbal third-person singular <byft.h> \textit{byftH} `he opens', with the nominal form <fy ft.h> \textit{fy ftH} `in opening'.
%
%
%

\begin{figure}[!h]
\begin{center}
\setlength{\tabcolsep}{3pt}
\begin{tabular}{llllll}
\hline
  {\bf Source:} & \multicolumn{5}{r}{<AlbAb> [[<fy ft.h>]] <AnA `ndy m^sklT>}\\
    & \textit{AnA} & \textit{{\AYN}ndy} & \textit{m{\SHIN}kl{\TAMARBUTA}} & [[\textit{fy ftH}]] &   \textit{AlbAb}\\
     & I &    to-me  & problem & [[in opening.N]]  &the-door\\
     \hline
  {\bf Output:} & \multicolumn{5}{r}{<AlbAb> [[<byft.h>]]  <AnA `ndy m^sklT>}\\
  & \textit{AnA} & \textit{{\AYN}ndy} & \textit{m{\SHIN}kl{\TAMARBUTA}} & [[\textit{byftH}]]  &\textit{AlbAb} \\
   &I & to-me & problem & [[with-opens.3MS]] & the-door\\
   \hline 
 {\bf Reference:} & \multicolumn{5}{l}{I'm having trouble opening the door} \\
 \hline 
\end{tabular}
\caption{Example \textsc{Dir+PP} error where the output does not preserve the POS of the source.}
\label{fig:ex1}
\end{center}
\end{figure}




Because Arabic verbs convey person in much finer granularity than do English verbs, which only inflect for third-person singular forms in present tense, many Arabic verb inflections in the source-to-pivot and pivot-to-target phrase tables will be aligned to the same morphologically deprived English verb, e.g., `opening'.
Thus, when the phrase tables are combined via shared English phrases, any given inflected Egyptian verb can be mapped to a large number of Levantine inflections, which, mostly, will not share the same morpho-syntactic properties. 
In this case, because `opening', like many `ing'-suffixed forms in English, can be nominal or verbal, it is not just inflectional morphology that is confused but derivational morphology, as the POS is misinterpreted.

\subsection{Addressing Resource Limitations}
\cite{kholy2015morphological} use AMEANA~\citep{ameana}, an automatic error analysis tool, to determine that definiteness, gender, and number are the features that most frequently contribute to such errors in Hebrew-to-MSA MT. In this work, we were not able to use AMEANA, as it relies on accurate morphological analyses that we cannot produce automatically for Levantine. Furthermore, even if we knew what were the most problematic features for translating Egyptian to Levantine, we might not be able to leverage them, as there is non-trivial noise and Egyptian bias in how the analyses were generated (Section \ref{preprocessing}). 

Our approach focuses instead on identifying features that: (i) MADAMIRA-EGY can recognize relatively accurately (ii) tend to be consistently translated from Egyptian to Levantine.
For instance, second-person and third-person verbal forms are frequently orthographically ambiguous in Arabic, making person challenging for our analyzer to correctly identify.
Thus, adding a constraint to the model promoting consistent translation of the person feature value would be useless because we are not likely to know the correct property of person in the first place. Furthermore, if the possible values of a given feature can be translated freely, modeling that feature will be similarly useless.
This is often the case with tense, as <b^swfk> \textit{b{\SHIN}wfk} `see you' in Egyptian could conceivably be translated as <r.h ^swfk> \textit{rH {\SHIN}wfk} in Levantine, changing the value from progressive (realized by the cliticized particle <b> \textit{b}) to future tense (realized by the particle <r.h> \textit{rH}).

Without gold, morphologically annotated data in Levantine, we cannot independently measure either our ability to identify morpho-syntactic feature values correctly, or the consistency with which they should be translated.
However, we can approximate both jointly.
Assuming that Egyptian feature values should correspond to the same feature values on the Levantine side,
we measure, for each morpho-syntactic feature, how frequently the realized property on the Egyptian side is aligned to the same property on the Levantine side. 

As shown in Table~\ref{consistencyTable}, definiteness, gender, number, and POS are the only features which map consistently across aligned tokens in more than 50\% of their occurrences throughout the BTEC training set.
This suggests that they both can be recognized accurately and are consistently preserved in human translation. Even so, the fact that none of these features map consistently over 80\% of the time, suggests that modeling such features will be noisy.

\begin{table}[!h]
\setlength{\tabcolsep}{2pt}
\centering
\begin{tabular}{|c|c|c|c|c|c|}
\hline
\textbf{Definiteness} & \textbf{Number} & \textbf{Gender} & \textbf{POS} & \textbf{Aspect} & \textbf{Person} \\ \hline
75                  & 75           & 62           & 56         & 32           & 29            \\ \hline
\end{tabular}
\caption{Percentage rates over all training set token alignments at which the feature's values were preserved from Egyptian to Levantine.}
\label{consistencyTable}
\end{table}

\subsection{Computing Constraint Scores}
Similar to \cite{kholy2015morphological}, we design morpho-syntactic constraints by calculating probability distributions from Egyptian to Levantine and vice versa.
These reflect how likely each morpho-syntactic property set on one side is to be aligned to each morpho-syntactic property set on the other, based on how often such alignments occured in the training data. 
Properties sets are defined as the conjunction of values for all morpho-syntactic features under consideration, which for us, include the four most "consistent" features as identified in Table \ref{consistencyTable}: definiteness, number, gender---which were used by \cite{kholy2015morphological}---and also POS (thus, masculine-singular-verb and definite-feminine-noun are property sets).
Unaligned tokens are considered to be aligned to a null token on the opposite side, 
and thus are mapped to an empty property set.
We use these probability distributions to add two constraint scores to every phrase pair in the phrase pivot table, one calculated from Egyptian to Levantine and the other, Levantine to Egyptian, as defined in Equations \ref{eq1} and \ref{eq2}. 

\begin{center}
\begin{equation} \label{eq1}
W_{s}  = \frac{1}{A} \sum\limits_{\forall (i,j) \in a} P(MLE(i) | MLE(j))  
\end{equation}
\end{center}

\begin{center}
\begin{equation} \label{eq2}
W_{t}  = \frac{1}{B} \sum\limits_{\forall (i,j) \in b} P(MLE(j) | MLE(i))  
\end{equation}
\end{center}

We calculate $W_{s}$ by summing over every alignment $a$ from source token $i$ to target token $j$ (if $i$ is unaligned, $j$ is the null token), the probability of $i$'s property set given $j$'s.
While \cite{kholy2015morphological} normalize this sum by the quantity of source tokens, we normalize by the total number of alignments $A$. Otherwise, many-to-one and one-to-many alignments would bias the scores and enable some to exceed one, making them impossible to interpret as probabilities.
The property sets of $i$ and $j$ are determined from the set of all possible property sets for that type (defined as the list of all unique analyses it received over every occurrence in BOLT, AOC, and the BTEC training data) via maximum likelihood estimation, $MLE$, so as to maximize the likelihood of the source property set given the target property set.

This entails that individual tokens' property sets can be analyzed differenlty from the source side than from the target side.
Also, sequences of $MLE$ property sets over multiple tokens on a single side can be syntactically infeasible, e.g., containing 5 consecutive verbs.
Thus, we experimented with additional constraints, requiring (i) aligned $MLE$ property sets to have been aligned at least once in the training set (ii) syntactic feasibility on source and target sides independently (iii) alignment of the sequence of property sets on the source side to that on the target side to have appeared at least once in the training set.
However, none of these experiments boosted performance, suggesting that $MLE$ inconsistency is not a problematic issue.

$W_{t}$ is calculated equivalently to $W_{s}$ from the target side. Adding these constraint weights to each phrase pair in the phrase pivot table, we re-tune the \textsc{Dir+PP} system, re-test, and obtain a statistically significant improvement with a score of 18.03 BLEU on the development set.

We then evaluate the \textsc{Direct}, \textsc{Dir+PP}, and Direct + Phrase Pivot with Morpho-syntactic Features (\textsc{Dir+PP+Morph}) systems on the 2,000 sentence blind test set from our BTEC corpus. The results in Table \ref{FinalResultsTable} confirm the utility of our added constraints, as each successive model significantly improves over the last, as in the development set.

\begin{table}[tbh]
\normalsize
\centering
\begin{tabular}{|c|c|c|c|c|}
\hline
\textbf{Model} & \textbf{Dev} & \textbf{Dev OOV} & \textbf{Test} & \textbf{Test OOV} \\ \hline
\textsc{No-Translation} & 6.48 & N/A & 6.45 & N/A \\ \hline
\textsc{Direct} & 15.44 & 4.6 & 14.61 & 5.4 \\ \hline
\textsc{Dir+PP} & 17.41 & 0.9 & 16.69 & 1.0 \\ \hline
\textbf{\textsc{Dir+PP+Morph}} & \textbf{18.03} & 0.9 & \textbf{17.48} & 1.0 \\ \hline
\end{tabular}
\caption{Comparing BLEU scores of systems with and without morpho-syntactic features on development and blind test sets. Out-of-vocabulary rates are reported as percentages.}
\label{FinalResultsTable}
\end{table}


\section{Error Analysis} \label{errorAnalysis}

We analyized the output of the \textsc{Dir+PP} and \textsc{Dir+PP+Morph} models on 100 development set sentences to investigate the effects of morpho-syntactic features and to identify issues for future work.
Table \ref{EA1} reveals a stark contrast---about a 30\% gap in both models---between the quantity of output tokens matching a reference token letter-for-letter (row 3), and the quantity of output tokens manually judged to be acceptable (row 2).
Approximately 10\% of that 30\% gap is due to lack of spelling standards in DA (row 4). \cite{Habash_12_LREC} developed a Conventional Orthography for Dialectal Arabic (CODA) to standardize spelling while preprocessing DA and a prototype system can CODA{\it fy} Egyptian \citep{Eskander-Codafy:2013}, though no such system is yet available for Levantine. Once developed, we expect such a system to improve MT quality not only by imposing consistent output, but also by reducing sparsity in all translation and language models during training.

The remaining 90\% of the 30\% gap between exactly matched tokens and tokens judged to be correct can be approximately split into thirds. One third cannot be directly linked to any reference token (row 7), e.g., tokens in paraphrasal or idiomatic constructions. Another third is tokens which can be linked to a reference token, but take a different root (row 5).
The final third are inflectional or derivational variants of the corresponding reference token (row 6).
The fact that so many inflectional/derivational variants
are judged correct demonstrates that morpho-syntactic modeling is necessarily noisy as property sets are frequently not preserved, even in acceptable translations.
On the other hand, the success of \textsc{Dir+PP+Morph} suggests that some features' properties tend to be preserved through translation or at least altered predictably, as otherwise, the system would not benefit from modeling them.

\begin{table*}[htb]
\centering
\setlength{\tabcolsep}{2pt}
\begin{tabular}{cc|c|c|}
\cline{3-4}
      \multicolumn{2}{c|}{}& \textbf{\begin{tabular}{p{1.5cm}}\textsc{Dir+PP}\end{tabular}} & \textbf{\begin{tabular}{p{1.5cm}}\textsc{Dir+PP}\\\textsc{Morph}\end{tabular}} \\ \cline{2-4}
            
1 & \multicolumn{1}{|l|}{\begin{tabular}{p{3.5cm}}\textbf{Words}\end{tabular}}         & 665    & 670           \\ \cline{2-4}

2 &\multicolumn{1}{|l|}{\textbf{Words Judged Correct}}    & \textbf{569 (85.6)}  & \textbf{594 (88.7)}         \\ \cline{3-4}

3 & \multicolumn{1}{|r|}{Exact Match}   & 377 (56.7)  & 382 (57.0)         \\ \cline{3-4}
4 &\multicolumn{1}{|r|}{CODA Variant} & 23 (3.5)  & 25 (3.7)         \\ \cline{3-4}

5 & \multicolumn{1}{|r|}{Different Root}    & 47 (7.1)  & 51 (7.6)        \\ \cline{3-4}
6 &\multicolumn{1}{|r|}{Different Properties}    & 61 (9.2)    & 65 (9.8)        \\ \cline{3-4}

7 &\multicolumn{1}{|r|}{Otherwise Different}    & 61 (9.2)    & 71 (10.7)        \\ \cline{2-4}\cline{2-4}

8 &\multicolumn{1}{|l|}{\textbf{Words Judged Incorrect}}          & \textbf{96 (14.4)}  & \textbf{76 (11.3)}         \\ \cline{3-4}
9 &\multicolumn{1}{|r|}{Morpho-syntactic Properties}   & 49 (7.4)   & 41 (6.1)         \\ \cline{3-4}
10 &\multicolumn{1}{|r|}{Other Problems}          & 47 (7.1)   & 35 (5.2)         \\ \cline{2-4}

11 &\multicolumn{1}{|r|}{\textbf{Properties}: Modeled}   & 33 (5.0)   & 26 (3.9)         \\ \cline{3-4}
12 &\multicolumn{1}{|r|}{Not Modeled}          & 16 (2.4)   & 15 (2.2) 
\\ \cline{2-4}

13 &\multicolumn{1}{|r|}{\textbf{Other}: Wrong Word Sense}   & 18 (2.7)   & 17 (2.5)         \\ \cline{3-4}
14 &\multicolumn{1}{|r|}{Apparent Phrasal Issue}          & 13 (2.0)   & 7 (1.0) 
\\ \cline{3-4}
15 &\multicolumn{1}{|r|}{Unclear Reason}          & 13 (2.0)   & 7 (1.0) 
\\ \cline{3-4}
16 &\multicolumn{1}{|r|}{OOV}          & 2 (0.3)   & 3 (0.4) 
\\ \cline{3-4}
17 &\multicolumn{1}{|r|}{Copies Egyptian}          & 1 (0.2)   & 1 (0.1) 

\\ \cline{2-4}\cline{2-4}

18 &\multicolumn{1}{|l|}{\begin{tabular}{p{3.9cm}}\textbf{Word Error Reduction}\end{tabular}}         & \textbf{N/A}  & \textbf{(20.2)}         \\ \cline{2-4}\cline{2-4}

19 &\multicolumn{1}{|l|}{\begin{tabular}{p{3.9cm}}\textbf{Sentences}\end{tabular}}          & 100    & 100           \\ \cline{2-4}
20 &\multicolumn{1}{|r|}{Correct Sentences}          & 48     & 55         \\ \cline{2-4}
21 &\multicolumn{1}{|r|}{Sentence Error Reduction}         & N/A    & (13.5)   \\ \cline{2-4}
\end{tabular}

\caption{Comprehensive manual error analysis of 100 sentences from the development set. Values within parentheses are percentages.}
\label{EA1}
\end{table*}

\begin{figure*}[!ht] 
\setlength{\tabcolsep}{3pt}
\footnotesize

\begin{tabularx}{\textwidth}{lllllllllll}

\hline

 \multirow{10}{*}{\normalsize{\bf (a)}} & \multicolumn{2}{l}{\bf ENGLISH:} & \multicolumn{8}{l}{Actually, inside it [...] Do you mind if I take it out?}\\
 
   & \multicolumn{2}{l}{\bf REFERENCE:} & \multicolumn{8}{r}{<`ndk mAn` AzA ^sltw?> [...] <bAl.hqyqT fy bAlbw>}\\

   & & & \textit{bAlHqyq{\TAMARBUTA}} & \textit{fy} & \textit{bAlbw} & [...] & \textit{{\AYN}ndk} & \textit{mAn{\AYN}} & \textit{AzA} & \textit{{\SHIN}ltw?}\\
   
   & & & with-the-truth & exists & in-heart-its & [...] & to-you & problem & if & take-1SPast-it \\
  
 & \multicolumn{2}{l}{\bf \textsc{Dir+PP}:} & \multicolumn{8}{r}{*<.tl`w?> <`ndk mAn` A_dA> [...] <fy bAlbA>  *<.hqyqT>}\\
 
  & & & \textit{Hqyq{\TAMARBUTA}*} & \textit{fy} & \textit{bAlbA} & [...] & \textit{{\AYN}ndk} & \textit{mAn{\AYN}} & \textit{A{\DHA}A} & \textit{Tl{\AYN}w?}*\\
  
   & & & truth* & exists & in-heart-its & [...] & to-you & problem & if & remove.3SPast-it*\\

 & \multicolumn{2}{l}{\bf \textsc{Dir+PP+MORPH}:} & 
    \multicolumn{8}{r}{<`ndk mAn` A_dA ^sltw?> [...] <Al.hqyqT, hwy bqlbw>}\\

  & & & \textit{AlHqyq{\TAMARBUTA}} & \textit{hwy} & \textit{bqlbw} & [...] & \textit{{\AYN}ndk} & \textit{mAn{\AYN}} & \textit{A{\DHA}A} & \textit{{\SHIN}ltw?}\\
   & & & the-truth & it & in-heart-its & [...] & to-you & problem & if & take.1SPast-it?\\
  \multicolumn{1}{l}{} \\ \hline \hline
  
  \end{tabularx}

\medskip


   
     
     


\medskip

\begin{tabularx}{\textwidth}{lllllllllllll}

    \multirow{10}{*}{\normalsize{\bf (b)}} & \multicolumn{2}{l}{\bf ENGLISH:} & \multicolumn{4}{l}{I'll bring one right away}\\
     
   & \multicolumn{2}{l}{\bf REFERENCE:} & \multicolumn{4}{r}{<r.h ^gyb wA.hd hlA>}\\
    & & & \textit{rH} & \textit{jyb} & \textit{wAHd} & \textit{hlA}\\
    & & & will & bring.1S & one & now\\
    
    & \multicolumn{2}{l}{\bf \textsc{Dir+PP}:} & \multicolumn{4}{r}{<wA.hd hlq> *<bAl^gybT>}\\
    & & & \textit{bAljyb{\TAMARBUTA}*} & \textit{wAHd} & \textit{hlq}\\
    & & & with-the-pocket* & one & now\\
    
     & \multicolumn{2}{l}{\bf \textsc{Dir+PP+Morph}:} & \multicolumn{4}{r}{<r.h ^gyb wA.hd hlq>}\\
    & & & \textit{rH} & \textit{jyb} & \textit{wAHd} & \textit{hlq}\\
    & & & will & bring.1S & one & now\\
     \multicolumn{1}{l}{} \\ \hline \hline
     
   \end{tabularx}

\medskip

\begin{tabularx}{\textwidth}{lllllllllllll}

     \multirow{10}{*}{\normalsize{\bf (c)}} & \multicolumn{2}{l}{\bf ENGLISH:} & \multicolumn{6}{l}{What kind of fruit do you have?}\\
     
     & \multicolumn{2}{l}{\bf REFERENCE:} & \multicolumn{6}{r}{<Ay nw` fwAky `ndk?>}\\
    & & & \textit{Ay} & \textit{nw{\AYN}} & \textit{fwAky} & \textit{{\AYN}ndk?}\\
    & & & which & kind & fruit & at-you\\
    
     & \multicolumn{2}{l}{\bf \textsc{Dir+PP}:} & \multicolumn{6}{r}{*<`ndk Ay nw` mn AlfAkhT ^sw?>}\\
    & & & \textit{{\AYN}ndk} & \textit{Ay} & \textit{nw{\AYN}} & \textit{mn} & \textit{AlfAkh{\TAMARBUTA}} & \textit{{\SHIN}w?*}\\
    & & & at-you & which & kind & from & the-fruit & what*\\
    
     & \multicolumn{2}{l}{\bf \textsc{Dir+PP+Morph}:} & \multicolumn{6}{r}{<`ndk Ay nw` mn AlfAkhT?>}\\
    & & & \textit{{\AYN}ndk} & \textit{Ay} & \textit{nw{\AYN}} & \textit{mn} & \textit{AlfAkh{\TAMARBUTA}?}\\
    & & & at-you & which & kind & from & the-fruit   \\ \hline 
  
\end{tabularx}

\caption{Example translation errors (marked with *) corrected by adding morpho-syntactic constraints to the model.}
\label{EAsents}

\end{figure*}

\subsection{Direct Effects of Added Features}
The second major insight of the error analysis is that the error reduction from adding morpho-syntactic constraints is far more significant (20.2\%, row 18), than the improvement registered by the automatic BLEU scores. Example sentences illustrating some of these improvements are contained in Figure~\ref{EAsents}.
For 7.4\% of the tokens \textsc{Dir+PP} outputs, its only mistake is misrepresenting one or more morpho-syntactic property (row 9).
Comparing that to 6.1\% for \textsc{Dir+PP+Morph} (row 9), the new model makes a 17\% error reduction in the area it is designed to improve.
Furthermore, essentially all of this improvement takes place in sentences where \textsc{Dir+PP+Morph} corrects a mistake involving a feature we model: definiteness, gender, number, or POS.
For example, the definite article is correctly added to the word <Al.hqyqT> \textit{AlHqyq{\TAMARBUTA}} `the truth' in Figure~\ref{EAsents}a.

\subsection{Indirect Effects of Added Features}
Surprisingly, most of the overall error reduction actually comes from mistakes other than misrepresentation of morpho-syntactic properties, as such mistakes decrease by 26\%, from 7.1\% to 5.2\% (row 10).
The morpho-syntactic constraints seem to teach the model about syntax at the phrase level, as sentences like Figure~\ref{EAsents}b are corrected, which were originally over-chunked into small phrases by \textsc{Dir+PP}.
<bAl^gybT> \textit{bAljyb{\TAMARBUTA}} `with the pocket' is likely a misanalysis of the source word <hAjyb> \textit{hAjyb} `I'll get' translated as a single-word phrase, as it would have made for an infrequent or non-existent bigram or trigram when combined with the following words in the output.
The \textsc{Dir+PP} model likely had access to longer phrase pairs such as <hAjyb wA.hd> \textit{hAjyb wAHd} `I'll get one' mapping to <r.h jyb wA.hd> \textit{rH jyb wAHd}---the corresponding reference phrase---but likely did not select it because longer phrases are inherently less frequent, i.e., noisier to model.

Morpho-syntactic constraints enable \textsc{Dir+PP+Morph} to increasingly select longer, infrequent phrase pairs by distinguishing those that are morpho-syntactically feasible from competing shorter alternatives. Translating larger phrasal chunks leads to more fluent output by reducing opportunities to incorrectly chunk phrase boundaries. This is why much of the error reduction does not appear to be, superficially at least, related to morpho-syntactic features targeted by \textsc{Dir+PP+Morph}. 

Figure~\ref{EAsents}c exhibits another benefit of morpho-syntactic features, as \textsc{Dir+PP+Morph} often corrects the insertion of gratuitous words, here, <^sw> \textit{{\SHIN}w} `what'. Such features teach the model that certain POS's are less likely to align to the null token, even if the language model favors the sequence with the gratuitous token monolingually.


\section{Conclusion and Future Work} \label{conclusion}


In this work, we presented the second ever evaluated Arabic DA-to-DA MT effort.
The subject has not previously received serious attention due to lack of naturally occurring parallel data, though DA is widely spoken and dialects are frequently mutually unintelligible, exhibiting comparable linguistic variation to the Romance languages.
Our results suggest that modeling morphology and syntax can significantly improve DA-to-DA MT despite data sparsity.
However, optimizing models under such circumstances requires careful consideration of the linguistic differences between dialects and careful tailoring and implementation of all available data and resources.

Given that many DA pairs may not have pivot data available, the most pressing future work is to develop a dialect-agnostic tokenizer and analyzer which does not suffer from the Egyptian bias that ours does.
This will reduce data sparsity regardless of the nature of the low-resourced MT settings for any DA pair, and it will enable better morpho-syntactic modeling.

Additionally, improving on the dialect identification work of \cite{Diab_10_LREC}, \cite{zaidan-callisonburch:2011:ACL-HLT2011}, and \cite{elfardy-diab:2013:Short} will enable us to collect more monolingual data.
This data is not only useful for language modeling but can also be mined for comparable sentences to augment the parallel training set.
The process typically involves using metadata \citep{resnik2003web} and seed data \citep{munteanu2005improving} to identify pairs of related sentences or phrases in the source and target languages \citep{cettolo2010mining, max2012generalizing}.
These are then iteratively classified via expectation maximization with phrases identified as parallel being added to the seed data \citep{dong2015iterative}.
Models trained thusly produce noisy phrase pairs, often imperfectly modeling morpho-syntactic property sets.
Thus, the same morpho-syntactic constraints that improved \textsc{Dir+PP+Morph} can be adapted to improve MT via comparable corpora. 

\section{Acknowledgments}
This publication was made possible by grant NPRP 7-290-1-047 from the Qatar National Research Fund (a member of Qatar Foundation). The statements made herein are solely the responsibility of the authors. The first author was funded by the Boren Fellowship program.


\end{document}